# FRONT MATTER

**Title**
An Autonomous Drone for Search and Rescue in Forests using Airborne Optical Sectioning

**Authors**
D.C. Schedl, I. Kurmi, and O. Bimber*

**Affiliations**
Computer Science Department, Johannes Kepler University Linz, Austria
*Corresponding Author. oliver.bimber@jku.at

## Abstract
Drones will play an essential role in human-machine teaming in future search and rescue (SAR) missions. We present a first prototype that finds people fully autonomously in densely occluded forests. In the course of 17 field experiments conducted over various forest types and under different flying conditions, our drone found 38 out of 42 hidden persons; average precision was 86% for predefined flight paths, while adaptive path planning (where potential findings are double-checked) increased confidence by 15%. Image processing, classification, and dynamic flight-path adaptation are computed on-board in real time and while flying. Our finding that deep-learning-based person classification is unaffected by sparse and error-prone sampling within one-dimensional synthetic apertures allows flights to be shortened and reduces recording requirements to one tenth of the number of images needed for sampling using two-dimensional synthetic apertures. The goal of our adaptive path planning is to find people as reliably and quickly as possible, which is essential in time-critical applications, such as SAR. Our drone enables SAR operations in remote areas without stable network coverage, as it transmits to the rescue team only classification results that indicate detections and can thus operate with intermittent minimal-bandwidth connections (e.g., by satellite). Once received, these results can be visually enhanced for interpretation on remote mobile devices.

## Summary
A prototype drone is presented that finds people in densely occluded forests fully autonomously.

# MAIN TEXT

## Introduction

Use of unmanned aerial vehicles (UAVs) by emergency services in civil applications, such as search and rescue (SAR), fire-fighting operations, policing, traffic control and disaster management, is rapidly increasing. In contrast to manned aircraft, such as helicopters, drones are more flexible and cheaper in acquisition, maintenance, and operation; moreover, they avoid risks to pilots under difficult weather conditions. Considerable progress is being made to create fully autonomous drones that team up with action forces. However, technical and legal obstacles remain.

Efficient path planning, for instance, is key in autonomous UAVs and poses serious challenges due to numerous constraints, such as limited energy, speed and payload, and vulnerability to various conditions. Previous studies have addressed these problems for single (*1-3*) or multiple cooperative (*4-6*) and non-cooperative (*7*) UAVs. They addressed relevant constraints, pursuing a variety of goals, such as trajectory optimization (*8,9*), path or motion planning (*2-3,10-13*), or vehicle routing (*14,15*). Various optimization techniques based on sampling (e.g., rapidly exploring random trees, probabilistic road maps, Voronoi diagrams, and artificial potential fields), graph-based algorithms (e.g., Dijkstra's algorithm, A*, D* and theta*), bio-inspired/evolutionary algorithms (e.g., ant colony optimization and particle swarm optimization), mathematical models (e.g., mixed-integer linear programming, satisfiability modulo theory, control theory, and heuristic approaches), and machine learning models (e.g., reinforcement learning and other artificial neural networks) have been used and are summarized in reviews such as (*2,16*). These approaches can also be grouped by approach, such as global path optimization (e.g., A*, particle swarm optimization, and satisfiability modulo theory, which are mainly static and computed offline) or local optimization (e.g., artificial potential fields and rapidly exploring random trees, which are mainly adaptive and computed in real time). Grid-based and artificial potential -field-based approaches have (due to their scalability, limited computational complexity, and adaptability) been widely used in the course of real-time adaptive path planning for obstacle avoidance (*17-19*) and target detection and tracking (*20,21*). Adaptive path planning has also been demonstrated for rescue operations within simulated indoor environments (*22*), where drone-person distance and viewing angle were optimized by means of reinforcement learning.

In addition to efficient path planning, effective imaging is essential for autonomous UAVs. The narrow aperture optics of conventional cameras increase the depth of field and therefore project sharply entire occlusion volumes (such as forests) into the images captured. Objects of interest at a particular distance often remain fully occluded. Wide aperture optics (i.e., with diameters of several meters), which cause an extremely shallow depth of field, would be better suited to these cases, but remain infeasible for airborne applications. Synthetic-aperture (SA) sensing offers a means of overcoming these physical limitations; it is widely recognized to approximate theoretical wide-aperture sensors by computationally combining the signals of multiple small-aperture sensors or a single moving small-aperture sensor to improve resolution, depth of field, frame rate, contrast, and signal-to-noise ratio. This principle has been applied in a range of fields, for instance, for obtaining weather-independent images and reconstructing geospatial depth information using radar (SAR) (*23-25*), for observing large celestial phenomena in outer space using radio telescopes (SART) (*26,27*), for reconstructing a defocus-free 3D volume using interferometric microscopy (ISAM) (*28*), and for applying 2D synthetic aperture imaging to shorter wavelengths (i.e., optical light) using LiDAR (SAL) / synthetic-aperture imaging laser (SAIL) (*29,30*). In the visible range, synthetic-aperture imaging (SAI) (*31-38*) has been used to acquire structured light fields (regularly sampled multiscopic scene representations). Systems employing large camera arrays capture 4D light-ray data which is then used to support various digital post-processing steps (such as refocusing, computation of virtual views with maximal synthetic apertures, and varying depth of field) after their acquisition. In such systems, synthetic apertures are constrained mainly by the physical size of the camera array used.

With Airborne Optical Sectioning (AOS) (*39-45*), we have introduced a wide synthetic-aperture imaging technique that employs manned or unmanned aircraft, such as drones (Fig. 1**A**), to sample images within large (synthetic aperture) areas from above occluded volumes, such as forests. Based on the poses of the aircraft during capturing, these images are computationally combined to integrate images. These integral images suppress strong occlusion and make visible targets that remain hidden in single recordings. As illustrated in Fig. 1**B**, a point on a given digital elevation model of the scanned environment's ground surface is projected into every sampling perspective on the synthetic aperture plane (the sampling plane of the drone). The corresponding pixel values from all projections are averaged. Repeating this for all surface points results in the integral image. In (*41*), we presented a statistical model to explain the efficiency of AOS with respect to occlusion density, occluder sizes, number of integrated samples, and size of the synthetic aperture. The main advantages of AOS over alternatives, such as LiDAR (*46-48*), are its computational performance in real-time occlusion removal and its applicability to other wavelengths, such as far infrared (thermal imaging) (*42*) for wildlife observations (*43*) or search and rescue (SAR) (*45*), or near infrared for agriculture and forestry applications.

In (*45*) we demonstrated that integrating single images before classification rather than combining classification results of single images is significantly more effective when classifying partially occluded persons in aerial thermal images. Here, AOS was applied for image integration and occlusion removal. This finding, in principle, enables drone-supported SAR missions in heavily concealed areas, such as forests. In practice, however, several challenges remain to be overcome: Thus far, drones have been used only to sample images along a predefined flight path that covers a two-dimensional synthetic aperture, with all image processing done offline, after a flight, on high-performance computers. Both sampling of large 2D aperture areas and heavy image processing (in particular computer-vision-based pose estimation, which can take hours even on high-end GPUs) require a significant amount of time, which makes them impractical for time-critical applications, such as SAR. Furthermore, sampling along a fixed flight path does not allow regional differences in occlusion density that may require re-sampling to be considered.

In this article, we report on two main contributions towards fully autonomous drones for search and rescue: First, we show that –compared to the classical approach of two-dimensional aperture sampling and computationally expensive computer-vision-based pose estimation– the effectiveness of person classification under occlusion conditions is not reduced when sampling along one-dimensional synthetic apertures (i.e., the 1D flight path) and using imprecise GPS and IMU measurements for pose estimation. Having resampled previous test flights (45), we report that imprecise 1D sampling of around 30 images and precise 2D sampling of around 300 images yield similar average precision scores of 92-93%. It follows that image processing can be implemented in its entirety using on-board mobile processors, thus enabling real-time classification during time-efficient flights. Second, we introduce an adaptive online sampling technique based on potential fields that changes the flight path dynamically during flight and uses classification confidences. Thus, the drone can decide to re-sample a particular region if it receives even a weak hint from the classifier that a person might be hidden there. We developed an autonomous prototype drone and evaluated it in the course of 17 field experiments over various forest types (conifer, broadleaf, mixed) and under different flying conditions (daylight, temperature, seasons). With this prototype, we achieved an average precision score of 86% for predefined flight paths and a 15% improvement in confidence scores for our adaptive path planning. From a total of 42 hidden persons, 38 were found by the drone (88% with one false alert for predefined flight paths, and 100% with no false alert with adaptive path planning).

# Results

## 1D Synthetic Aperture Classification

In (*45*), we sampled two-dimensional synthetic-aperture (SA) areas of 30 m × 30 m (this equals the ground coverage of our thermal camera's field of view at an altitude of approx. 35 m above ground level) with a sampling density of 1 m × 3 m. The high number of samples (approx. 300 images) and precise computer-vision-based pose estimation ensured effective occlusion removal and a focussed appearance of persons in the corresponding integral images (Fig. 2**A**,**B**). A one-dimensional SA sampling (i.e., integrating images captured along a one-dimensional flight path) and image registration with instant but imprecise GPS and IMU measurements from the drone results in defocus (Fig. 2**C**). The defocused point-spread of the thermal signal of a person contains the forest's occlusion structure in its optical bokeh.

The advantage of 1D over 2D SA sampling is a significant reduction in sampling time for covering the same area (by a factor of 10 in our example, as –at the same sampling rate– only 30 rather than 300 images are captured and integrated). The disadvantages are less efficient occlusion removal (due to undersampling) and defocus (due to less precise pose estimation). In particular, the latter has implications not only for classification itself, but also for training a classifier. In (*45*) we showed that, in the case of efficient occlusion removal and precise image registration, the data needed for training a classifier is invariant to occlusion. Hence, training data can be recorded under control conditions in an open-field environment with the trained classifier still performing well for occluded test data recorded from above all types of forest. This is no longer the case for defocused 1D SA samplings, as the occlusion structure becomes an apparent part of each integral image's optical bokeh. Occlusion invariance can no longer be assumed, and, accordingly, training data must be produced in the presence of occlusion. However, the question arises of whether classification remains effective for cases such as that shown in Fig. 2**C** even if training data contains occlusion.

An initial experiment provided first evidence that this is the case: We split 11 test flights (F0-F11, excluding F7) over various forest types (conifer, broadleaf, and mixed forest) from (*45*) into five training flights (F0, F4, F5, F10, and F11, with 132 labels of 19 persons), two validation flights (F1 and F8, with 65 labels of 10 persons), and four test flights (F2, F3, F6, and F9, with 204 labels of 26 persons). Details of these flights are provided in (*45*). Note that, due to its circular sampling, we did not use F7. The original 2D grid SA sampling of each flight was computationally resampled to sequences of 1D line SA samplings, while the measured GPS/IMU data was used for pose estimation in place of a computer-vision-based pose estimation. Furthermore, a simpler classifier (YOLOv4-tiny (*49*)) that is suitable for less performant mobile processors was employed (previously, we applied the computationally more expensive YOLO3 classifier with spatial pyramid pooling (*50,51*)). Results are summarized in Fig. 3 and presented in more detail in Tab. 1. Fig. 3 plots the overall (of all test flights combined) average precision score (average precision, AP, is used to evaluate the performance of object classification algorithms and considers correct, wrong, and missed detections across multiple scenes (*52*)) over an increasing number N of integrated thermal images sampled along a 1D SA path. Samples were taken at intervals of 1 m. A maximum path length of 30 m (30 samples) was recorded in our flights. The results of discrete path-length variations were fitted to a hyperbolic function (see Sect. *Average Precision Curve* of the supplementary material) and extrapolated. This function indicates that high classification rates can be achieved for relatively small numbers of samples and short flight paths -- even for 1D SA sampling with defocus and simpler classifiers. For example, an average precision of 92.8% was achieved with N=30). In comparison, for the 2D SA sampling with better focus and classifier, an average precision of 92.2% was achieved with N>300 (*45*). Note that we augmented the training data of our new classifier with all path-length variations (from N=1 to 30). This finding suggests that neither 2D SA sampling nor precise pose estimation is necessary for achieving similar classification performance at significantly higher speed. However, training data is no longer invariant to occlusion. Fig. 3 also illustrates that the steep increase in classification performance followed by a flattening beyond a small N is in line with the increase in visibility improvement in integral images for increasing N. This was confirmed by our statistical model described in (*41*), where we claimed that the performance of synthetic aperture imaging for occlusion removal does not increase infinitely with increasing apertures and higher sampling rates, but that there is a (relatively low) limit in both for achieving a maximum improvement. These low limits make AOS practical for time-critical applications, such as SAR.

**Predefined Search**
In the second series of experiments, we validated our findings in practice in the course of 8 new test flights (covering a total area of approx. 6 ha) over various forest types, where all processing was carried out in real time on the drone and during flight, but where the flight paths were still predefined and static (following fixed waypoints). Here, we used test flights from (*45*) for training (F0, F2-F6, F8, F10, and F11) and training validation (F1 and F9), and, again, augmented with path-length variations from N=1 to 30. During each new test flight, the drone sampled segments of 30 m long 1D SAs with 1 sample/meter along defined waypoints. Thus, after each segment, an integral image was computed and classified. Tab. 2 presents the results, while Fig. 4. illustrates them for the example of flight F16. Of 34 persons present, 30 persons were found, and only one person was incorrectly detected. The overall average precision score was lower than in the first experiments (86% vs. 93%). This is due to Tab. 1 comparing test flights of our previous offline-classification study presented in (45), where compass errors were corrected manually after recording. The data of the new test flights in Tab. 2 were computed fully automatically during flights and contain compass errors. Better compass modules would be required to improve the results.

Many real-time object-detection algorithms that are based on convolutional neural networks (CNN), such as YOLO (*53*), divide images into regions, for each of which they predict bounding boxes and probabilities (confidence scores). These confidence scores are thresholded for making a final classification decision. Depending on the choice of threshold, either too many false positives (incorrect detections) or too few true positives (i.e., persons are missed) can be the result. For both series of experiments (Tab. 1 and Tab. 2), we selected a YOLO confidence score threshold of 10%. This minimizes false positives. A lower confidence score would increase the number of incorrectly detected persons (false alerts), but might also increase correct detections.

Since confidence scores correlate directly with the amount of occlusion, areas with detections that indicate the presence of a person should be re-sampled (double-checked) to observe the development of the confidence score before making a final decision. If an initially low score drops further after re-sampling, this confirms a false positive detection. If an initially high score increases, this confirms a true positive detection. However, this is only possible if the flight path is adapted dynamically in flight on the basis of current classification results.

**Adaptive Search**
In the third series of experiments, we dynamically computed and adapted the flight path on-board the drone rather than have it follow a predefined set of fixed waypoints. Here, the goal was to find a person as quickly and as reliably as possible, allowing the drone to make the decisions autonomously on how and where to search. Since the drone's flight path changes dynamically based on classification results, we chose to build on a potential-field based path-planning approach.

The search areas were split into grids of 30 m × 30 m cells, and each cell was initialized with the rescue team's estimate of the likelihood of finding a person within it.

Each cell was scanned centered (horizontally or vertically) from edge to edge with a 30 m SA, as in all previous experiments. The decision on which cells are scanned and in which order was based on the cell probabilities and by evaluating the following potential-field equation (as explained in (1)) for each cell:

$$f(i) = P(i) * e^{-\|x - c_i\|_2}, \quad (1)$$

where $P(i)$ is the $i$th cell's probability, $x$ is the current location of the drone, and $c_i$ is the center position of cell $i$. The cell scanned next was that with the highest potential $f(i)$. If multiple cells had the same maximum $f(i)$, we selected (among them) the cell $j$ neighboring $i$ that had the highest potential density (successively enlarging the neighborhood until a maximum potential density cell was identified):

$$P_d(i) = \sum_{j \neq i} \frac{P(j)}{\|c_i - c_j\|_2}. \quad (2)$$

The probabilities of scanned cells were set to zero to avoid revisiting them. We stopped scanning under three conditions: (i) if a confirmed true positive was detected; (ii) if a maximum path length (maximum flying time) was reached; (iii) if the entire search region had been covered. Fig. 5**A** illustrates an example path that was computed dynamically for a search area with an initial probability map. By evaluating Eqns. 1 and 2, we maximized the likelihood of detection while minimizing search time based on the rescue team's initial assessment of the situation.

If a detection was made (we considered a detection to be weak if it had a confidence score of at least 5%), re-sampling the corresponding cell was triggered. Since the orientation of occluders is unknown, we re-sampled with an SA that was orthogonal to the previous one to increase the likelihood of visibility. This new SA was not cell-centered, but centered on the detection. For multiple detections within the same cell, we scanned multiple SAs accordingly. Fig. 5**B** presents an example showing the sub-area highlighted in Fig. 5A in which two detections were made at one border of the cell (with confidence scores of 11% and 27%). After re-sampling, an increase in the confidence score of one detection (from 27% to 51%) confirmed a true positive, while a drop in the confidence score of the other detection (from 11% to 0%) confirmed a false positive. Fig. 5**C** illustrates another example, where a weak 7% confidence detection was re-sampled and, based on a 0% confidence score, confirmed to be a false positive.

Tab. 3 summarizes the results of 9 adaptive-search test flights. On average, re-sampling increased the confidence scores by around 15% for true positives (persons found) and decreased confidence scores by around 16% for false positives (persons incorrectly found). With a 10% confidence threshold as used before, and without adaptive re-sampling, false decisions would have been made in 4 of the 9 flights: 2 persons would not have been detected and 2 erroneous detections would have occurred.

## Discussion

For automatic person classification in occluding forests, 1D synthetic-aperture imaging performs – despite GPS errors– equally well as 2D synthetic-aperture imaging of 10 times the number of samples combined with time-consuming precise pose estimation. Nevertheless, image integration remains necessary to achieve effective classification rates. This finding motivated the development of a first fully autonomous SAR drone that operates adaptively and in real time. Unlike video streaming and remote-processing, on-board processing does not require fast and stable network coverage, which might not always be available for SAR missions in remote areas. Intermittent minimal-bandwidth satellite connections are sufficient to transmit classification results of detections only.

The transmitted integral images of detections are, due to GPS pose-error-related misregistration, not easy to interpret visually. Automatic pose-error reduction can be applied in the course of post-processing to achieve significant improvements, as explained in (*54*). These improvements, however, will not be beneficial to the automatic person classification itself, as small regions of interest around the findings are required for visual optimization, which are the result of the classification. Registration optimization for entire images is not feasible, as processing time and error-proneness would be too high. Thus, visual enhancement of received classification results can be computed on remote mobile devices to support the rescue team in decision-making. Figure S4 of the supplementary material illustrates the results of offline registration enhancement for all correct person detections (PF) in Tab. 3.

Our prototype and field experiments are clearly limited. Due to current aviation regulations, all test and training flights required visual line of sight. Short battery life restricted flying times to a maximum of 15-20 minutes. Both constrained the size of the test sites to 0.3-4.0 hectares only. Professional drones with combustion engines (e.g., boxer or wankel) are more suitable for SAR missions, as they can fly for up to 6 hours and carry payloads of up to 75 kg. Appropriate legal regulations for autonomous flights beyond visual line of sight are in development and will most likely be approved much earlier for emergency operations (search and rescue, fire fighting, disaster management) than for other applications. The drop in classification rate in our experiments from 93% to 86% that was due to remaining compass errors could be avoided by using more precise compass modules. The thermal camera we employed supports capture rates of 1 image/second. For one sample per meter, this restricted flying speed to 1 m/s and integral image computation and classification to discrete 30 m steps (fast flight segments without sampling were done with 3 m/s). Faster thermal cameras would enable higher flying speeds and continuous integral image updating and classification after capturing each image (e.g., after every meter flown). Potential-field-based path planning makes local decisions and does not lead to globally optimal solutions. A good trade-off between real-time dynamic adaptability and considering larger neighborhoods for path planning should be investigated. Furthermore, multi-drone solutions could reduce search time even further. All of these improvements, including enhancement of occlusion removal and classification, and application to other fields (such as wildlife observation, discovering fire hot spots, surveillance, and border control), fall within the scope of our future work. We believe that autonomous drones such as ours will play an essential role in human-machine teaming of future rescue missions and emergency operations.

## Materials and Methods

We equipped an octocopter (MikroKopter OktoXL 6S12; 945 mm diameter; approx. 4.5 kg; two LiPo 4500 mAh batteries) with a thermal camera (FLIR Vue Pro; 9 mm fixed focal length lens; 7.5 µm to 13.5 µm spectral band; 14 bit non-radiometric; 118 g), a single-board system-on-chip computer (SoCC) (RaspberryPi 4B; 5.6 cm × 8.6 cm; 65 g; 8 GB ram), an LTE communication hat (Sixfab 3G/4G & LTE base hat and a SIM card; 5.7 cm × 6.5 cm; 35 g), and a Vision Processing Unit (VPU) (Intel Neural Compute Stick 2; 7.2 cm × 2.7 cm × 1.4 cm; 30 g). The equipment (total weight 320 g) was mounted on a rotatable gimbal, and the camera pointed downwards during flight, as shown in Fig. 1.

The SoCC established communication with the drone (receiving IMU/GPS positions and sending waypoint instructions including GPS location, orientation, and speed) via a serial protocol, triggered the thermal camera using a pulse-width modulated (PWM) signal and its purpose input-output pins (GPIO), downloaded the recorded images from the camera's memory, preprocessed the images, and computed the integral image during flight. Preprocessing involved adjusting the mean of the thermal images, removing the lens distortion using OpenCV's pinhole camera model, and cropping the images to a field of view of 50.82° and a resolution of 512 px × 512 px. Preprocessing and integration of 30 thermal images required around 748 ms and 90 ms on the SoCC. The VPU was initialized with our pre-trained network weights for object detection using YOLOv4-tiny (49) at startup and detected persons in the integral images while flying. Detections were computed in 84 ms per integral image and were transmitted via an LTE connection to a remote mobile device using the communication hat. The main software running on the SoCC was implemented using Python, while submodules, such as the drone communication protocol and the integral computation, were implemented using C, OpenGL, and C++ and integrated using Cython. How integral images are computed was explained in detail in (*39,42,45*). This work differs in that we did not assume the ground surface to be planar, but approximated it with a digital elevation model (DEM). The DEMs for the test sites are available freely from the Upper Austrian State Government (*55*) in a common GeoTIFF format. We converted a predefined region to a triangle mesh using the GDAL software. The central position of each SA was used to compute the integral images, within which persons were labeled if visible.

For training and validation datasets, the following augmentations were applied: We randomly rotated the integral images (original and 9 random rotations), changed the altitude of the DEM from -3 to +3 m in steps of 1 m (7 variations), and varied the number of image samples per SA (pinhole, 1 to 5, 5 to 10, 10 to 15, 15 to 20, 20 to 25, and 25 to 30), thus obtaining a total of 490 augmentations per SA. Note that 6 out of 115 individual SAs were shorter than 25 cameras, and one SA was shorter than 20 cameras, which resulted in fewer long SAs.

For training our object detector with the YOLOv4-tiny (*49*) network architecture, we classified a single object class (person) in 512 px × 512 px integral images. The initial 29 network layers (out of 39 layers) were pre-trained with the Imagenet (*56*) dataset. We used the Darknet software for training and set the starting learning rate and the batch and subdivision sizes to 0.00261, 64, and 16, respectively. Training weights were stored after every 200 iterations and evaluated on the training set. The weights with the highest AP score (IoU of 25%) were used in the subsequent experiments (Tabs. 1, 2, and 3). The false positive (FP), true positive (TP), and average precision (AP) scores in Tab. 1 were computed in Matlab with an IoU threshold of 25%. For the FP and TP scores, detections below 10% confidence were discarded.

For our experiments summarized in Tab. 1, we reused the dataset from (*45*) and computationally resampled the recordings to sequences of 1D line SAs using the original GPS/IMU measurements. To make the new results comparable to the old 2D SA results in (*45*), we retained the manual compass correction as described in (*45*). For all other experiments, a manual compass correction was not possible, as images and drone poses were processed immediately in real time during flight. For the experiments in Tab. 1, we split 11 test flights (F0-F11, excluding F7) from (*45*) into five training flights (F0, F4, F5, F10, and F11), two validation flights (F1 and F8), and four test flights (F2, F3, F6, and F9). The classifier for the additional test flights F12 to F28 (in Tabs. 2 and 3) was trained using flights F0 to F11 (excluding F1, F7, and F8) for training and F1 and F8 for validation. For the experiments summarized in Tab. 2, the drone followed a pre-planned path, recorded images, computed integral images, and performed detections after every 30 m SA segment during flight. After a flight, persons were labeled and AP, persons found (PF), and persons incorrectly found (PI) scores computed. For accurate labeling, we asked subjects to record their GPS locations on site using their smartphones. Due to the overlapping fields of view, the same person might appear in multiple integral images. This is considered in the PF and PI metric in Tab. 2. For PF we considered a person as found if at least one detected bounding box (with at least 10% confidence score) overlapped with the person's GT label. Because of the GPS/IMU pose errors, we applied an IoU threshold of 1% and ignored multiple detections in the same GT bounding box. We classified a detection as incorrect (PI) if it had a confidence score above 10% but was outside of all GT bounding boxes.

For the experiments summarized in Tab. 3, the potential-field equations (Eqns. 1,2) were evaluated on the SoCC to decide on the next target cells based on the drone's current position, the probability map (predefined and updated after visiting each cell), and previous detection result. Computation time for planning the next target cell was 1 ms.

# Acknowledgments

**Funding:** This research was funded by the Austrian Science Fund (FWF) under grant number P 32185-NBL, and by the State of Upper Austria and the Austrian Federal Ministry of Education, Science and Research via the LIT – Linz Institute of Technology under grant number LIT-2019-8-SEE-114. **Author Contributions:** O.B. is the PI and originated the concept. O.B. and D.S. conceived and designed the experiments. D.S. and I.K. performed the experiments. O.B., D.S., and I.K. analyzed the data. D.S. and I.K. contributed materials/analysis tools. O.B., D.S., and I.K. wrote the paper. **Competing interests**: None. **Data and material availability:** The data collected in flights F12 to F28 can be downloaded from Zenodo (*74*) and includes ground-truth labels, single images, integral images for evaluation, network configuration files, trained network weights, and classification results. Flights F0 to F11 are available at (*75*). We provide labels and integral images for the resampled 1D apertures (*74*). Scripts to compute Tables 1 and 2 are provided with the dataset. **Code availability:** The full source code of the software is freely available on GitHub at https://github.com/JKU-ICG/AOS/.

# Figures and Tables

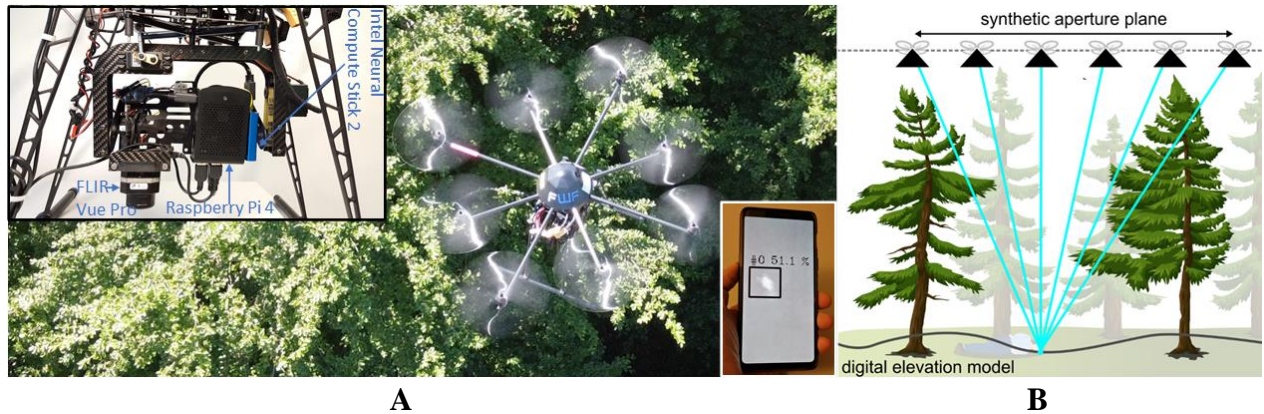

**Figure 1. Autonomous drone supporting search and rescue missions.** (**A**) Camera drone with the recording and processing equipment used in our experiments. In the case of a positive finding, an integral image with classification result and location is transmitted to a mobile device of the rescue team. (**B**) Basic principle of airborne optical sectioning (AOS). Supplementary Movie S1 shows video footage of drone and sample test site.

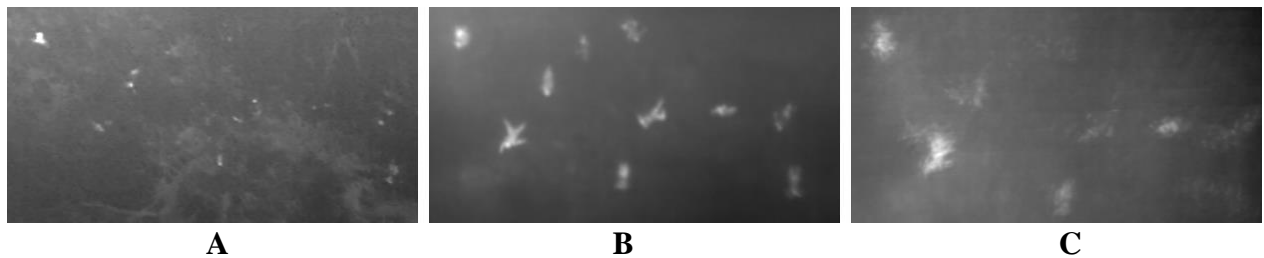

**Figure 2. 2D vs. 1D synthetic-aperture imaging.** The strong occlusion in single thermal images (**A**) can be removed effectively in integral images that densely sample two-dimensional synthetic aperture (**B**). Precise computer-vision-based pose estimation leads to well-defined shapes of people. One-dimensional synthetic aperture sampling combined with GPS/IMU-based pose estimation leads to defocus in the resulting integral images (**C**).

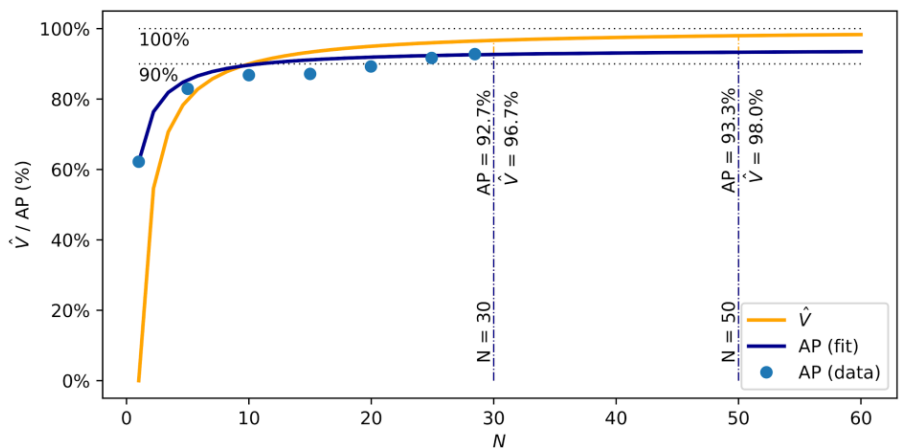

**Figure 3. 1D synthetic-aperture classification performance.** Classification performance increases for all test flights in Tab. 1 (blue) and modeled visibility improves (*41*) (yellow) with increasing number of integrated sample images N. Supplementary Movie S2 illustrates the behavior of the blue curve for integral images of 4 sample flights.

| flight (GT) | N=1 (pinhole) | | | N=5 | | | N=10 | | | N=15 | | | N=19.97 | | | N=24.93 | | | N=28.46 | | |
| --- | --- | --- | --- | --- | --- | --- | --- | --- | --- | --- | --- | --- | --- | --- | --- | --- | --- | --- | --- | --- | --- |
| | AP | TP | FP | AP | TP | FP | AP | TP | FP | AP | TP | FP | AP | TP | FP | AP | TP | FP | AP | TP* | FP* |
| F2 (70-86) | 64% | 65 | 80 | 90% | 78 | 16 | 94% | 81 | 8 | 96% | 80 | 12 | 97% | 81 | 7 | 96% | 81 | 5 | 97% | 66 | 5 |
| F3 (53) | 36% | 24 | 42 | 46% | 30 | 31 | 52% | 31 | 24 | 55% | 32 | 21 | 60% | 37 | 19 | 72% | 40 | 19 | 74% | 43 | 28 |
| F6 (65) | 86% | 56 | 16 | 99% | 64 | 1 | 99% | 64 | 4 | 98% | 64 | 2 | 98% | 64 | 1 | 98% | 64 | 1 | 98% | 64 | 2 |
| F9 (0) | n/a | 0 | 6 | n/a | 0 | 7 | n/a | 0 | 2 | n/a | 0 | 4 | n/a | 0 | 2 | n/a | 0 | 2 | n/a | 0 | 2 |
| ALL (188-204) | 62% | 145 | 144 | 83% | 172 | 55 | 87% | 176 | 38 | 87% | 176 | 39 | 89% | 182 | 29 | 92% | 185 | 27 | 93% | 173 | 37 |

**Table 1. 1D synthetic-aperture classification performance.** Classification results of test flights from (*45*), resampled to 1D SA segments, for various path lengths N. GT is the number of ground-truth labels, TP are the true positives, FP are the false positives, and AP is the average precision score. F9 is a flight over an empty (no people) forest. Note that, due to resampling, the path lengths per flight may vary slightly. Thus, N represents the average. Note also that fewer segments could be resampled for F2 (N=28.46) than for the other flights. This also led to fewer ground-truth labels (70 instead of 86) in this case. Movie S1 of the supplementary material visualizes the classification improvement for increasing N. Figure S1 of the supplementary material provides satellite and aerial RGB samples of the corresponding test sites.

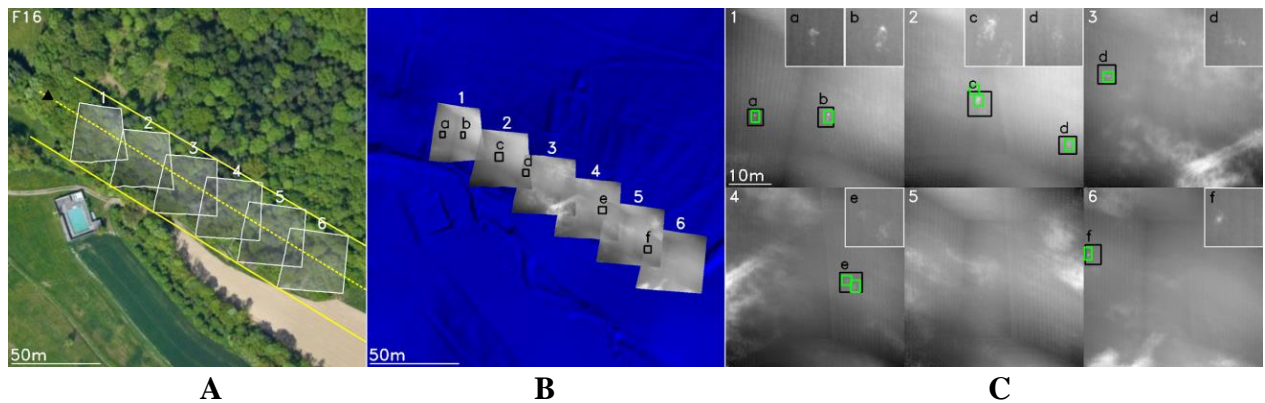

**Figure 4. Predefined search.** Visualization of the F16 test flight from Tab. 2 (see also Supplementary Movie S3). The flight path was predefined (dotted yellow line in **A**). The field of view (solid yellow lines in **A**) along the path was split into 30 m segments (plus residue). An integral image was computed for each segment (thermal image overlays in **A,B**). **A** shows a satellite view and **B** the digital elevation model of the test site. **C** shows the integral images and close-ups with locations at which persons are present (black boxes) and classifications results where persons are detected (green boxes). Note that the drone's orientation was kept constant (facing north) to reduce dynamic compass errors during flight.

| ID | latitude | longitude | date | path length | forest type | AP | PP | PF | PI |
|---|---|---|---|---|---|---|---|---|---|
| F12 | 48.33279428 | 14.33015898 | 13 Oct 20 | 119.4m | conifer | 96.7% | 2 | 2 | 1 |
| F13 | 48.33295299 | 14.33105885 | 13 Oct 20 | 263.9m | broadleaf | 100.0% | 2 | 2 | 0 |
| F14 | 48.33292935 | 14.33053121 | 15 Oct 20 | 388.0m | conifer, broadleaf | 100.0% | 6 | 6 | 0 |
| F15 | 48.33292935 | 14.33053121 | 20 Oct 20 | 394.8m | conifer, broadleaf | 75.6% | 6 | 5 | 0 |
| F16 | 48.33980546 | 14.33186291 | 22 Oct 20 | 182.3m | mixed | 91.1% | 6 | 6 | 0 |
| F17 | 48.33980546 | 14.33186291 | 22 Oct 20 | 180.0m | mixed | 63.3% | 6 | 4 | 0 |
| F18 | 48.3400508 | 14.33236699 | 28 Oct 20 | 181.8m | mixed | 100.0% | 2 | 2 | 0 |
| F19 | 48.3400508 | 14.33236699 | 28 Oct 20 | 211.9m | mixed | 75.0% | 4 | 3 | 0 |
| ALL | | | | 1922.1m | | 86.2% | 34 | 30 | 1 |

**Table 2. Predefined search experiments.** Results of test flights with predefined flight paths, including GPS coordinates, date of flight, path length, and forest type at test sites. AP is the average precision score, PP the number of persons present at the test site, PF the number of persons correctly found, and PI the number of incorrect person detections (false alerts). Note that flight paths were divided into 30 m segments (plus residues). Note also that, due to overlapping fields of view of the integral images, the same person may be labeled multiple times. Since it matters whether – and not how many times – a person was detected during the flight, we report person-specific PP, PF, and PI figures rather than label-specific GT, TP, and FP values as in Tab. 1. The results of F16 are shown in Fig. 4**C**, and the results of all other flights are illustrated in Figure S2 of the supplementary material. Aerial RGB samples of the corresponding test sites are shown in Figure S1 of the supplementary material.

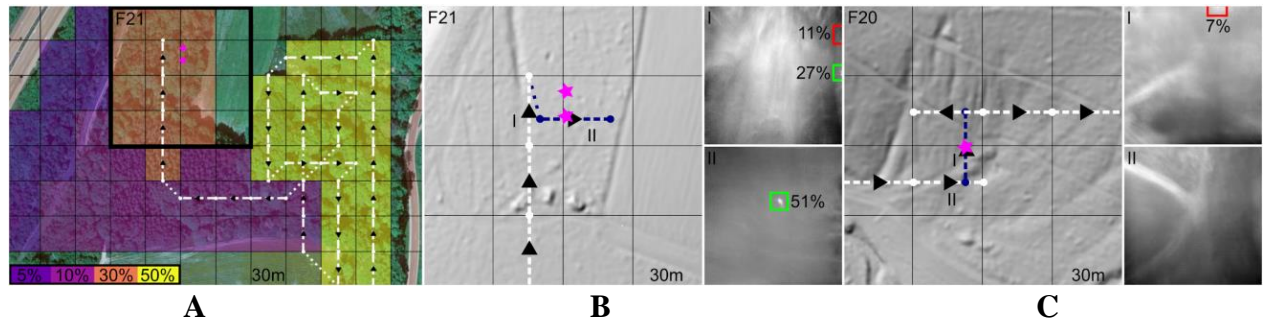

**Figure 5. Adaptive search.** Satellite image of the test site overlaid with the flight path computed dynamically based on potential fields (**A**). The initially provided probabilities are color coded. Dashed white lines indicate scanned SAs within $30 \times 30$ m cells (1 m/s). Dotted white lines indicate fast flight segments (3 m/s) without sampling. Black triangles, the orientations of which illustrates the flight direction, mark positions of computed integral images. See also Supplementary Movie S4. Digital elevation model overlaid with the close-up sub-area of flight F21 (**B**). The two detections (pink stars) made in the initial integral image I are confirmed to be one correct (green box) and one incorrect (red box) classification in the re-sampled integral image II. Re-sampled SAs and fast flight segments without sampling are indicated by dashed and dotted black lines, respectively. The case of an empty scene in which an incorrect detection was confirmed (F20 in Tab. 3) is shown in **C**.

| ID | latitude | longitude | date | forest type | PF ΔC( C ) | PI ΔC( C ) |
|---|---|---|---|---|---|---|
| F20 | 48.3398254 | 14.3327864 | 27 Nov 20 | mixed | empty | -6.7%( 0.0% ) |
| F21 | 48.3338445 | 14.3301653 | 30 Nov 20 | mixed | 23.9%( 51.1% ) | **-10.4%**( 0.4% ) |
| F22 | 48.3338445 | 14.3301653 | 30 Nov 20 | mixed | 8.4%( 44.2% ) | none |
| F23 | 48.3327007 | 14.330069 | 01 Dec 20 | conifer | 15.5%( 38.7% ) | **-40.4%**( 0.1% ) |
| F24 | 48.3332141 | 14.3314393 | 01 Dec 20 | broadleaf | **16.7%**( 22.6% ) | none |
| F25 | 48.3332141 | 14.3314393 | 02 Dec 20 | broadleaf | 2.6%( 35.6% ) | none |
| F26 | 48.3327007 | 14.330069 | 03 Dec 20 | conifer | **19.6%**( 19.6% ) | -6.9%( 0.0% ) |
| F27 | 48.3327007 | 14.330069 | 04 Dec 20 | conifer | 18.2%( 39.5% ) | none |
| F28 | 48.3338445 | 14.3301653 | 04 Dec 20 | mixed | 11.3%( 43.6% ) | none |
| ALL | | | | | 14.5% | -16.1% |

**Table 3. Adaptive-search experiments.** Results of test flights with adaptive flight paths, including GPS coordinates, date of flight, path length, and forest type at test sites. ΔC is the change in confidence score from the initial sampling of a cell to re-sampling of the corresponding detections for persons found (PF) and persons incorrectly found (PI). (C) is the final confidence score after re-sampling. Cases highlighted in bold would have led to misclassifications (persons not found or false alerts) for a confidence threshold of 10%. Fig. 5**B,C** shows cases F20 and F21, and Figure S3 of the supplementary material illustrates all other cases. Figure S1 of the supplementary material provides satellite and aerial RGB samples of the corresponding test sites.

# SUPPLEMENTARY MATERIALS

**Movie S1. Autonomous drone supporting search and rescue missions.** Video footage of our prototype drone and sample test site (**D** in Fig. S1).

**Movie S2. 1D synthetic-aperture classification performance.** Video footage illustrates the improvement in classification performance with increasing number of samples (N) for 4 sample test flights (broadleaf, conifer, mixed forest, and empty broadleaf forest). Shown are ground-truth labels (black boxes), true positives (green boxes), and false positives (red boxes).

**Movie S3. Predefined search.** Video footage illustrating test flight F16 (Fig. 4). The white pyramid indicates the field of view and the camera positions of the predefined flight path. Computed integral images with classification results are shown (green boxes: persons correctly found). Close-ups show the locations of persons in the integral image.

**Movie S4. Adaptive search.** Computed flight paths (without re-sampling in the case of detections) for various potential fields (probability maps), ranging from uniform probabilities to smooth and cluttered probability distributions, and the examples shown in Fig. 5**A**. The initially provided probabilities are color-coded. For visited cells, the color coding is removed. Dashed white lines indicate scanned SAs within 30 m × 30 m cells. Dotted white lines indicate fast flight segments without sampling. Black triangles, the orientations of which illustrate the flight direction, mark positions of computed integral images.

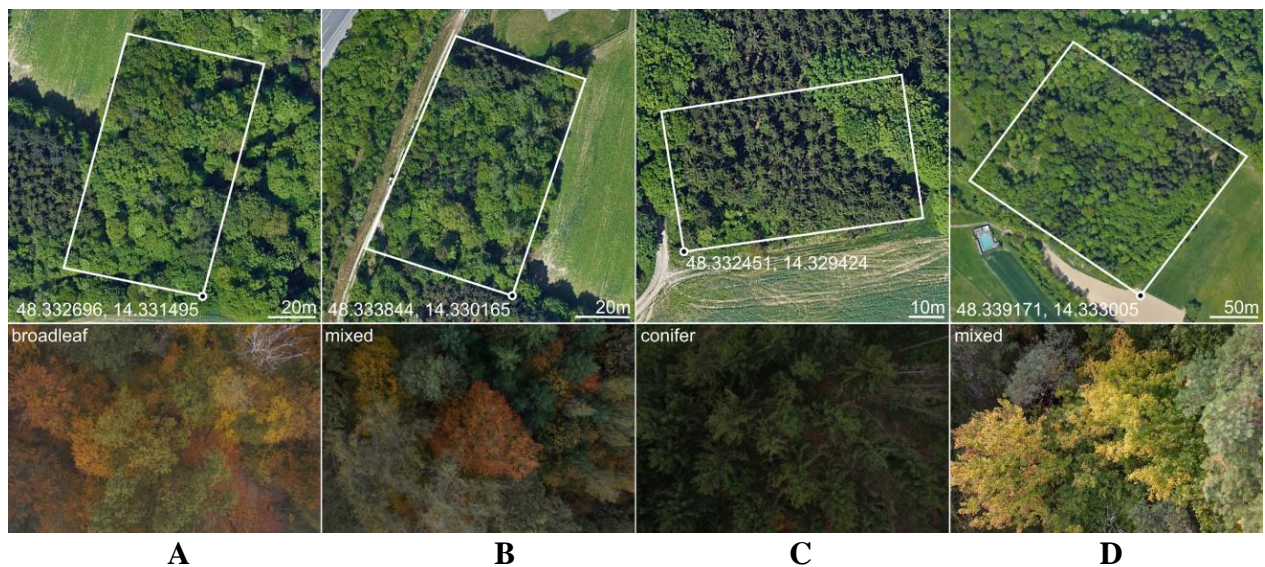

     **A**            **B**            **C**            **D**

**Figure S1. Test sites.** Satellite images and example drone (RGB) images of the test sites: Broadleaf forest site for test flights F2, F9, F13, F14, F15, F24, and F25 (**A**). Mixed forest site for test flights F3, F21, F22, and F28 (**B**). Conifer forest site for test flights F6, F12, F14, F15, F23, F26, and F27 (**C**). Second mixed forest site for test flights F16, F17, F18, F19, and F20 (**D**). Note that flights F14 and F15 covered both the conifer and the broadleaf sites.

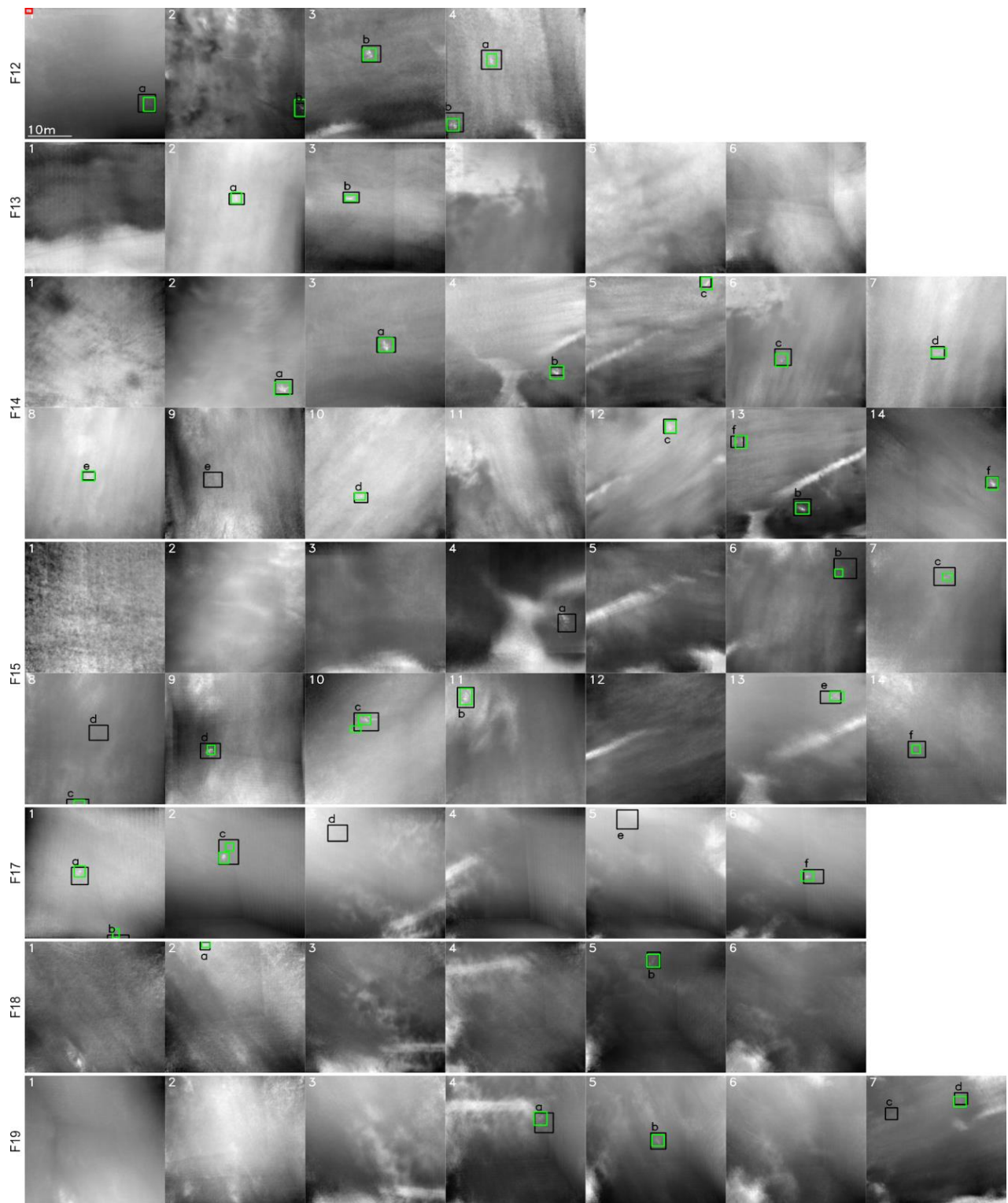

**Figure S2. Predefined search experiments.** Integral images and classification results for the test flights presented in Tab. 2 (except F16, which is shown in Fig. 4**C**). Black boxes indicate where persons are present, and green boxes are classified as containing persons. The red box in F12 indicates a false classification.

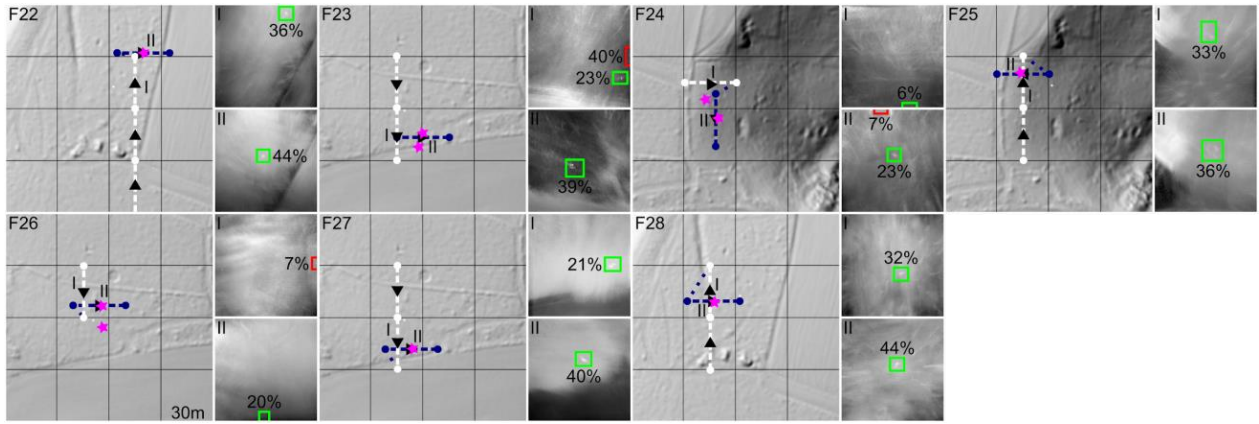

**Figure S3. Adaptive Search Experiment.** Remaining detection and re-sampling cases from Tab. 3 (except F20 and F21, which are shown in Fig. 5**B,C**). Digital elevation models overlaid with sub-areas. Pink stars indicate detections made. They are confirmed to be correct (green box) or incorrect (red box) classifications in the resampled integral images II. Re-sampled SAs and fast flight segments without sampling are indicated by dashed and dotted black lines, respectively. Black triangles, the orientations of which illustrate the flight direction, mark positions of computed integral images.

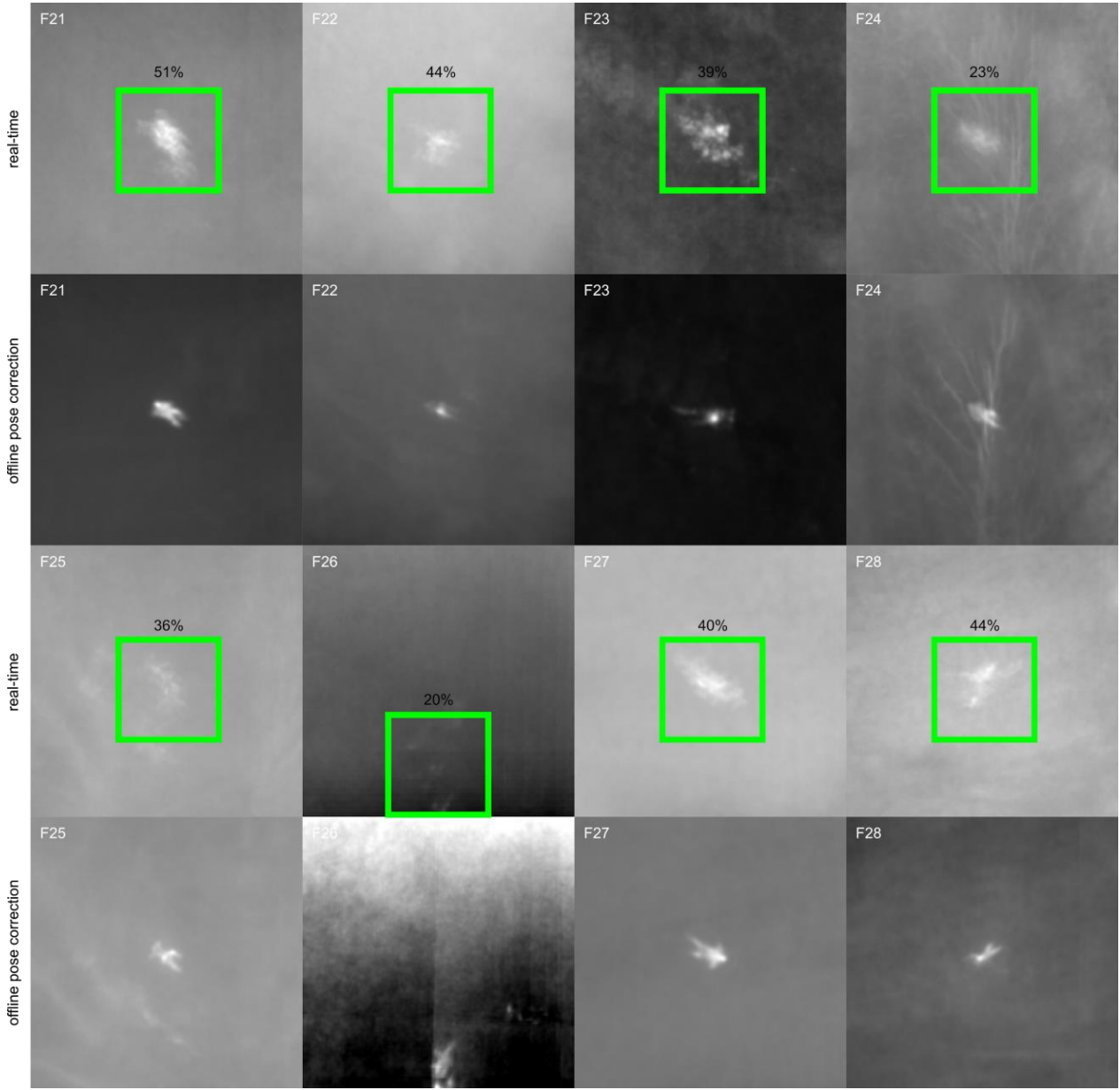

**Figure S4. Offline registration enhancement.** Results of the offline registration enhancement for all correct people detections (PF) in Tab. 3. Integral image computed in real time from GPS and IMU data on the drone (odd rows). Despite substantial misregistrations, person classification was successful. After automatic registration enhancement of the selected region of interest (bounding box of the classification result) as explained in (*54*), visual image quality is improved (even rows). Computation time was approx. 2-3 minutes on a standard PC with Nvidia GTX 1060 GPU, for 30-39 integrated samples and a region of interest with a resolution of 60 px × 60px.

**Average Precision Curve**

The AP (fit) curve in Fig. 3 shows a fit of the average precision (*AP*) scores over all scenes with *N* as a parameter (data in Tab. 1). The curve is a hyperbolic function of the form

$$AP(N) = \frac{a\,N}{b+N}, \quad (3)$$

with parameters *a* = *0.943* and *b* = *0.516*. The parameters were computed with a least-squares optimizer using the SciPy Python library, and we penalized positive residuals (estimations below the data points). The mean-squared error between the fitted curve and the data points in Tab. 1 is 0.0005.